\documentclass{article}
\usepackage{FairnessInML}

\usepackage{times}
\usepackage{soul}
\usepackage{url}
\usepackage[hidelinks]{hyperref}
\usepackage[utf8]{inputenc}
\usepackage[small]{caption}
\usepackage{graphicx}
\usepackage{amsmath}
\usepackage{amsthm}
\usepackage{booktabs}
\usepackage{algorithm}
\usepackage{algorithmic}
\usepackage{subcaption}
\usepackage{multicol}
\newcommand{\indep}{\perp \!\!\! \perp}

\title{Fairness-Aware Learning with Prejudice Free Representations}

 \author{
     Ramanujam Madhavan \and Mohit Wadhwa
     \affiliations
     Artificial Intelligence, LinkedIn
     \emails
     rmadhavan@linkedin.com \and mwadhwa@linkedin.com
 }

\begin{document}

\maketitle

\begin{abstract}
  Machine learning models are extensively being used to make decisions that have a significant impact on human life. These models are trained over historical data that may contain information about sensitive attributes such as race, sex, religion, etc. The presence of such sensitive attributes can impact certain population subgroups unfairly. It is straightforward to remove sensitive features from the data; however, a model could pick up prejudice from latent sensitive attributes that may exist in the training data. This has led to the growing apprehension about the fairness of the employed models. In this paper, we propose a novel algorithm that can effectively identify and treat latent discriminating features. The approach is agnostic of the learning algorithm and generalizes well for classification as well as regression tasks. It can also be used as a key aid in proving that the model is free of discrimination towards regulatory compliance if the need arises. The approach helps to collect discrimination-free features that would improve the model performance while ensuring the fairness of the model. The experimental results from our evaluations on publicly available real-world datasets show a near-ideal fairness measurement in comparison to other methods.
\end{abstract}

\section{Introduction}
Machine learning models are increasingly being employed to aid decision-makers in all kinds of consequential decision settings such as healthcare, education, employment, criminal justice, etc. This amplifying use of model-based decision-making process has raised concerns from policymakers, legal experts, philosophers, and civil organizations about the prejudice (also called as bias) in the employed decision models that rely on the statistical patterns learned from the real-world data ~\cite{datta2015automated,sweeney2013discrimination,zhang2019faht,zliobaite2015survey}.

A machine learning algorithm learns from historical instances (i.e., training data) to produce decisions for future instances. The learning algorithms are designed to learn statistical patterns in the training data, and since these decision systems use real-world data, the unintended consequence is that it can also potentially learn any unfair prejudice that may exist in the real-world data. For example, credit scoring is often decided based on the past credit data records, and the past credit data records may contain less sampled minority classes that could result in random predictions for minority classes. Another well-known example is of ProPublica COMPAS's risk assessments -- COMPAS algorithm was being used to forecast which criminals are most likely to re-offend -- where it was found that the results were displayed differently for black and white offenders ~\cite{angwin2016machine}.

As the application of machine learning and artificial intelligence is becoming more prevalent, there is an increasing concern to make the decision process socially and legally fair i.e. non-discriminatory and fair in sensitive traits such as race, sex, religion, etc.  Several techniques have been proposed to reduce or eliminate prejudice in machine learning models. A naïve approach is to avoid using sensitive features such as race, sex, religion, etc. while learning the model. However, machine learning algorithms are capable of learning complex relations present in the data, and there could be certain indirect data points that may serve as the proxy for these sensitive features. For example, the mention of the phrase “Convener of Grace Hopper Summit” in the resume may possibly indicate that the candidate is a female though there is no explicit mention of the sex. Hence, we need a more holistic mechanism to ensure that the data is free of such potential unfair prejudice inducing features. 

To address these challenges, we propose an algorithm that automatically detects and treats the prejudice inducing features from the dataset and then learns a fair model. The process to detect and treat prejudice inducing features here is interpretable and easily verifiable. Our approach processes the dataset in an iterative way and removes information about the features which are latently related to the known sensitive attributes. The processed prejudice-free features can be used by the downstream tasks to build fair models. The information about the detected prejudice inducing features can also be used to understand the discrimination in the dataset and the dataset can thus be altered accordingly. By providing interpretable information about the prejudice inducing features in the dataset, our technique allows to collect better prejudice-free features that are more directly related to the model target variable than sensitive attributes. To validate our approach, we do a detailed experimental evaluation on real-world datasets that show significant improvement in fairness measures over prevailing methods. We have shown the comparison of our technique against other well-known methods ~\cite{calders2010three,feldman2015certifying,kamishima2012fairness,zafar2017fairness}.

The remainder of this paper is organized as follows. Section 2 defines the basic notations. Section 3 reviews the related work. Section 4 discusses the proposed approach to eliminate prejudice inducing features from the dataset and learn a fair model. The experimental evaluation results are presented in Section 5. Finally, we conclude the paper in Section 6.  

\section {Basic Notations}
Unfair treatment of people based on sensitive attributes such as race or sex is prohibited by anti-discrimination laws in many countries. Disparate impact is one of the metrics to evaluate fairness under these laws. Disparate impact occurs when decision system outcomes disproportionately impact certain sub-populations. While there is no mathematical formula for quantifying disparate impact, we use 80\% rule (or more generally p\% rule) advocated by the US Equal Employment Opportunity Commission (EEOC) ~\cite{feldman2015certifying}. Disparate impact is calculated as the proportion of the unprivileged group that received the positive outcome by the proportion of the privileged group that received the positive outcome i.e. 
\\

$\textit{Disparate Impact} = \frac{Pr(\hat{Y}=1 | S=unprivileged)}{Pr(\hat{Y}=1 | S=privileged)}$
\\
\\
where $\hat{Y}$ represent the predicited variable and $S$ represent the sensitive attribute class.

For cases with more than two possible values of sensitive attributes, two variants of disparate impact could be considered i.e. binary and average. In the binary case, all unprivileged and privileged classes are grouped together into one unprivileged and privileged class respectively, and then a standard disparate impact formula is applied. In average case, all pairwise disparate impact calculations are done for each of the unprivileged and privileged class combinations, and the average over all these calculations is reported ~\cite{friedler2019comparative}. 

A disparate impact score of near $1.0$ would mean that the model is fair. A lower value would mean that the model is unfair to the unprivileged population, and a value above $1.0$ would mean that the model favors the unprivileged population. We use disparate impact as a fairness measure to demonstrate and validate the performance of our algorithm. 

Disparate treatment is another notion used by anti-discrimination laws to evaluate the fairness of a decision-making system. This notion occurs when known sensitive information is used by the model to make decisions. We avoid disparate treatment by ensuring that known sensitive features are removed from the dataset before learning the model ~\cite{zafar2017fairness}.

Performance measurement for a binary classification problem at various thresholds settings is represented by AUC-ROC (Area Under the Curve - Receiver Operating Characteristics) curve which is the primary metric we use to make dataset prejudice free. ROC is a probability curve, and AUC represents the degree of measure of separability. Higher AUC score conveys that the model is more capable of differentiating between target (positive and negative) classes. An AUC score of 0.5 conveys that the model has no capacity to distinguish between positive and negative target class. We leverage the AUC metric to make the dataset indifferent to target classes for each of the known sensitive features ~\cite{bradley1997use}. In the case of multi-class classification problems, one vs. all techniques could be used for calculating the AUC metric.

We show accuracy vs. disparate-impact trade-off for which accuracy of classification prediction is calculated as:
\\

$\textit{Accuracy} = \frac{\text{Number of correct predictions}}{\text{Total number of predictions}}$ 
\\
\\
For the purpose of this paper, we restrict the explanation and demonstration scope of our technique to binary classification tasks, but the technique can be easily extended to multi-class classification and regression tasks. 

\section {Related Work}
Fairness approaches generally fall into one of the following categories: pre-processing, in-processing, and post-processing. In each of these categories, a number of approaches have been proposed to address prejudice and learn fair machine learning models.  

Pre-processing approaches try to learn a new representation by removing correlation to (latent) sensitive attributes and preserves the training dataset information as much as possible. These new representations can then be fed to downstream tasks to build a model free of prejudice. Some advantages of these are that they don't modify the existing learning algorithm implementation, and access to the sensitive attributes is not required during the inference time. The technique we propose in this paper falls under this category of fairness approaches. One of the well-known previous work in this category is by ~\cite{zemel2013learning} that tries to learn a new representation of the data that is independent of the protected attribute by formulating fairness as an optimization problem while retaining as much information as possible about the features to encode the data and simultaneously obfuscating any information about membership in the protected group. ~\cite{kamiran2009classifying} is another well-known work in this category that selects the data points which are closer to the decision boundary and ranks them. The label of the ranked data points are carefully swapped in order to maintain the balance between classes and to minimize the loss of predictive accuracy.

In-processing approaches take algorithms into consideration, contrary to pre-processing approaches which are algorithm agnostic, and modifies the cost function to also account for fairness objectives. The most common idea leveraged by these approaches is to add a penalty or a regularization term to the existing optimization objective. ~\cite{calders2010three} proposes three naïve-bayes classifier approaches that fall under this category. The first approach modifies the probability of being positive to alter the decision distribution, the second one trains a model per sensitive attribute and balances them, and the third one optimizes the latent variable representing the prejudice free target by using expectation maximization methods. Another method is ~\cite{kamishima2012fairness} that modifies the cost function to add a regularization term in order to restrict the model learner’s behavior from sensitive information. 

Post-processing approaches, on the other hand, try to correct trained model predictor in terms of fairness constraints by adjusting the decision regions. In ~\cite{hardt2016equality}, the operating threshold of the machine learning model is decided based on the ROC curves with respect to the targeted groups. The point of intersection of these ROC signify equalized odds i.e. both true and false positive rates are equal for targeted groups. The point where true positives rates are equal for targeted groups represents fairness according to the equality of opportunity notion. 

\section {Prejudice Free Representations }
As in the case of any supervised learning setting, we assume that we have access to a labeled dataset (X, S, Y) where X denotes feature vectors, S denotes known sensitive features, and Y represents the target variable. We also assume that the domain of each of the columns in S is limited to discrete values, which is a common scenario in most of the real-world tasks, for example, sex, race as sensitive features. We also assume that known sensitive attributes are already removed from X, thereby avoiding disparate treatment i.e. features in S and X are mutually exclusive. The technique we propose here effectively identify and remove latent discriminating features present in the dataset (X, S, Y), while learning the model to predict target variable Y. 

\subsection{Problem Formulation}
For the dataset represented by $(X, S, Y)$, let $\mathcal{M}(Y|X, S)$ be a prediction model that models the conditional distribution of $Y$ given $X$ and $S$. Our objective is to ensure that model $\mathcal{M}$ is free of prejudice from sensitive features $S$ and latent sensitive features among $X$. In other words, the conditions $Y \indep  S$ $\vert$  $X$ and $S \indep X$ should be satisfied to make the model prejudice free. The first condition is easy to achieve by simply eliminating the sensitive features S from prediction model $\mathcal{M}$, however $Y \indep S$ $\vert$  $ X$ doesn’t guarantee $Y \indep S$: for example one or more sensitive features are linear of combinations of known non-sensitive features: $S = X + \epsilon_s $. Hence, we need a method to remove prejudice by making X and Y independent from S simultaneously \cite{kamishima2012fairness}.

Before defining a general framework, we pick logistic regression as an example classification algorithm to discuss the approach that ensures $Y \indep S$ and $S \indep X$ conditions. Let $D = \{(x, s, y)\}$ be the dataset comprising of the instances of random variables $X$, $S$ and $Y$. We model $\mathcal{M}(Y|X, S; \theta)$ to predict the conditional probability of target variable Y where $\theta$ represents the set of model parameter that are learnt. In a typical case, the model parameters $\theta$, are tuned to maximize the log-likelihood given by equation 1:
\begin{equation}
\resizebox{.91\linewidth}{!}{$
	\displaystyle
\begin{split}
	\hat{\theta} & = \arg\max_{\theta} \big[ \ln \mathcal{M}\big(Y|X, S; \theta\big) \big] \\
&  = \arg\max_{\theta} \big[\sum_{(x, y) \in D} \big(y\ln{\frac{1}{1 + e^{-x^{T}\theta}}} + \big(1 - y\big)\ln{\frac{1}{1 + e^{x^{T}\theta}}}\big) \big]
\end{split}
$}
\end{equation}
In our modified setting, we learn the prejudice free model in a two-step process through equation 2 and 3 where $D' = \{(x, s, y)\}$ be the dataset comprising of the instances of random variables $X'$, $S$ and $Y$:
\begin{equation}
\resizebox{.99\linewidth}{!}{$
	\displaystyle
\hat{X'} = \arg\min_{X^{'} \subset X} \big[ \arg\max_{\omega} \big[ \sum_{(x, s) \in D'} \big(s\ln{\frac{1}{1 + e^{-x^{T}\omega}}} + \big(1 - s\big)\ln{\frac{1}{1 + e^{x^{T}\omega}}}\big) \big] \big]
$}
\end{equation}%

\begin{equation}
\resizebox{.91\linewidth}{!}{$
	\displaystyle
	\hat{\theta'} = \arg\max_{\theta'} \big[\sum_{(x, y) \in D'} \big(y\ln{\frac{1}{1 + e^{-x^{T}\theta'}}} + \big(1 - y\big)\ln{\frac{1}{1 + e^{x^{T}\theta'}}}\big) \big]
	$}
\end{equation}%

Equation 2 minimizes the maximum likelihood of the model that learns $\omega$ to predict the sensitive variables $S$ from $X’$ where $X'$ is a subset of features from $X$ after eliminating the most predictive features in an iterative manner. Equation 3 learns the maximum likelihood parameter to predict $Y$ based on  prejudice free features $\hat{X}$ which is the residual set of $X$, and therefore ensures the condition $Y \indep S$. The Prejudice Free Representations (PFR) algorithm in the next section provides a general framework for this approach that is agnostic of the learning algorithm.

\subsection{General Framework}
In supervised learning, models evolve over iterations of feature engineering and hyperparameter tuning to optimize for one or more key performance metrics such as precision, recall, ROC, AUC, etc. In feature engineering, we tend to add more features to enhance the model’s predictive ability.  Our approach here applies what we call the inverse of feature engineering where we remove features iteratively so that the AUC of the model that predicts the sensitive variable falls until the point when the model has no capacity to distinguish between the classes of the sensitive variable.

\begin{figure*}[tb]
	\begin{multicols}{3}
		\begin{subfigure}[b]{1.0\linewidth}
			\includegraphics[width=\linewidth]{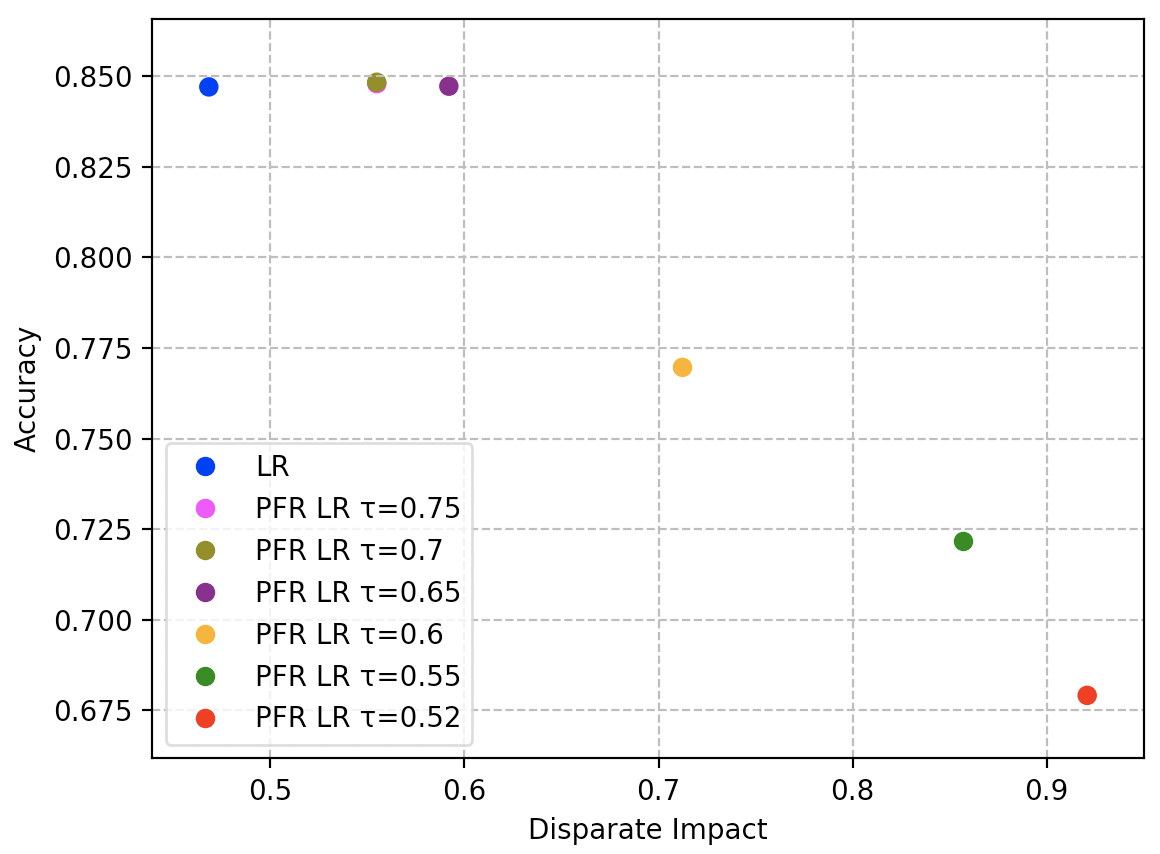}
			\caption{Race Sensitive Attribute }
		\end{subfigure}\par
		\begin{subfigure}[b]{1.0\linewidth}
			\includegraphics[width=\linewidth]{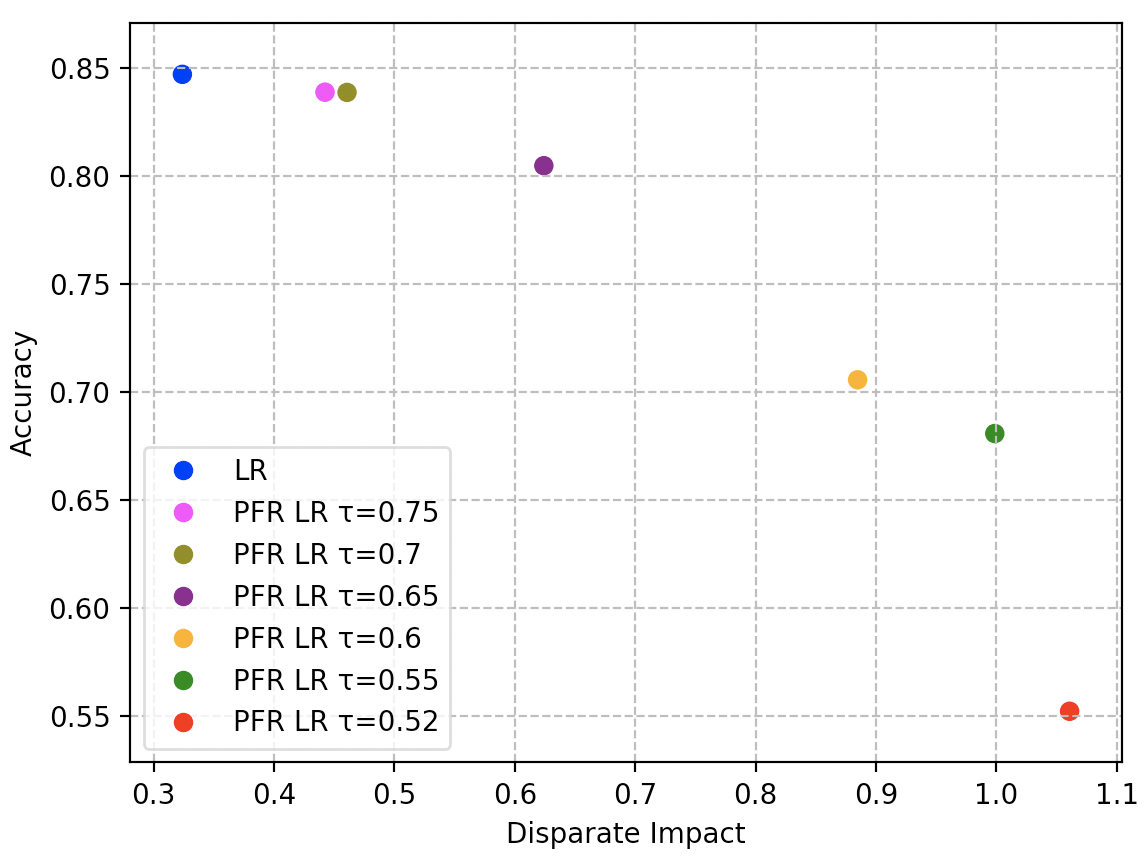}
			\caption{Sex Sensitive Attribute }
		\end{subfigure}\par
		\begin{subfigure}[b]{1.0\linewidth}
			\includegraphics[width=\linewidth]{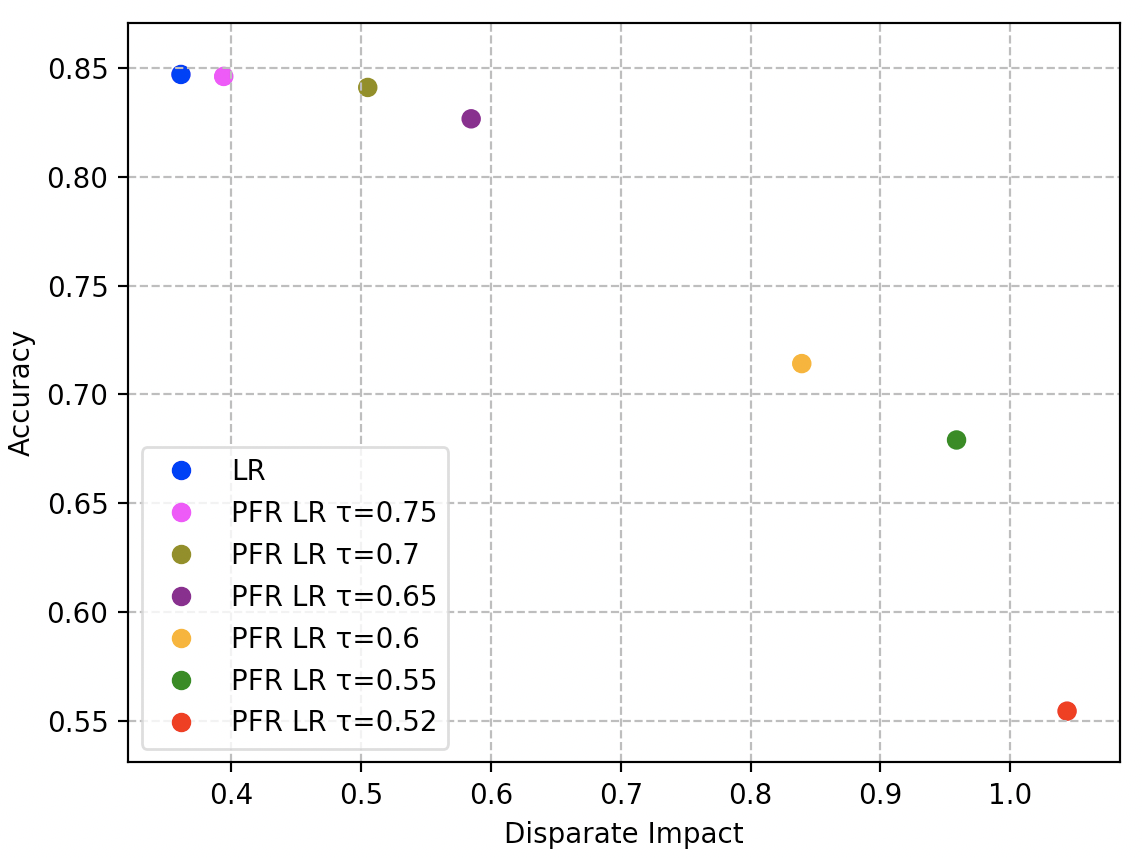}
			\caption{Race \& Sex Sensitive Attributes }
		\end{subfigure}
	\end{multicols}
	\caption{Accuracy vs. Disparate Impact evaluation on Adult Income Dataset}
	\label{fig:aidpfr}
\end{figure*}

\begin{algorithm}[tb]
	\caption{Prejudice Free Representations}
	\label{alg:algorithm}
	\textbf{Notations}: 
	\\ $\zeta$ -\  function to train a classifier 
	\\ $\psi$ -\ function to get AUC score
	\\ $\eta$ -\ function to get most important feature
	\\
	\textbf{Input}: 
	\\ $X$ -\ feature vectors
	\\ $S$ -\ known sensitive features
	\\ $\tau$ -\ auc thresholds for sensitive features
	\\
	\textbf{Initliaze}: $X^\prime$ = $\frac{X -min(X)}{max(X) - min(X)}$ \quad \quad // min-max scaling $X$\\
	\textbf{Output}: prejudice free dataset \\
	\textbf{Process:}
	\begin{algorithmic}[1] 
		\STATE $\forall (s_i, \tau_i) \in (S, \tau)$
		\STATE \quad \quad $\mathcal{M}$ := $\zeta(X^\prime, s_i)$ 
		\STATE \quad \quad $auc$ := $\psi(\mathcal{M}, X^\prime, s_i)$
		\STATE \quad \quad \textbf{while} $auc > \tau_i$ \textbf{do}
		\STATE \quad \quad \quad \quad $f$ := $\eta(\mathcal{M}, X^\prime)$
		\STATE \quad \quad \quad \quad $X^\prime$ := $X^\prime \setminus f$ 
		\STATE \quad \quad \quad \quad $\mathcal{M}$ := $\zeta(X^\prime, s_i)$
		\STATE \quad \quad \quad \quad $auc$ := $\psi(\mathcal{M}, X^\prime, s_i)$
		\STATE \quad \quad \textbf{end while}
		\STATE \textbf{return} $X^\prime$
	\end{algorithmic}
\end{algorithm}

Prejudice Free Representations (PFR) algorithm identifies the latent discriminating features and removes them from the dataset. To do so, it iteratively trains a classifier for each of the features in $S$ as target variable and removes the most important feature from the feature set, X, until the AUC of the sensitive variable predicting model drops to the acceptable fairness measure threshold represented by $\tau$. The process is repeated for each of the features in $S$.  While the underlying learning algorithm attempts to maximize AUC, the PFR algorithm drives the AUC down by removing most predictive feature in every iteration. In this process, latent features that have a correlation to the sensitive features are removed in a thorough manner. The output of the algorithm can be fed to the learning algorithm to learn prejudice free prediction model for target variable Y. Taking an example Logistic Regression (LR) classifier, the PFR algorithm outputs a prejudice free dataset as per our modified setting in equation 2, and the output dataset is then used to train a LR classifier as per equation 3. For the LR classifier, the most important features is the one with the maximum absolute weight where the features are normalised.

Algorithm 1 above provides a pseudocode implementation of the PFR algorithm. The algorithm provides a (best-first) greedy feature selection and removal mechanism to eliminate the prejudice inducing features, but the formulation and approach could be generalized to other feature selection methods as well. A few things to take note of are that the feature vectors of $X$ are scaled before applying the algorithm as it requires relative feature importance values which can be known only when the features are normalized. The other thing is about the AUC threshold value $\tau$. The value $\tau$ for a binary classifier is calculated as a function of target class imbalance. For example, if the domain of target variable is $\{0, 1\}$, then the $\tau_i$ for a sensitive feature $s_i$ is calculated as:
\begin{equation}
\resizebox{.41\linewidth}{!}{$
	\displaystyle
	\tau _i= \frac{\max(\vert s_i \setminus \{0\} \vert, \vert s_i \setminus \{1\} \vert)}{\vert s_i \vert}
	$}
\end{equation}%
It is therefore advisable to set AUC threshold $\tau$ as per equation 4. The parameter $\tau$ could be varied to handle Accuracy vs. Fairness-Measure trade-off as discussed more in the experimental section below.

The algorithm generalizes well for any classification algorithm where the feature importance score could be known, such as logistic regression, decision tree-based classifiers, neural networks, etc. As we use the same algorithm that is used to train the target model to identify and remove the prejudice inducing features, the approach provides a verifiable guarantee that the target model is free of prejudice. The dataset processed by the algorithm can also be fed to downstream learning tasks such as regression and classification to build prejudice free models. For example, the prejudice free datasets created using logistic regression algorithm can be used to train a linear regression, or an SVM model.  

\section {Experiments}
We evaluate our approach on a couple of publicly available real-world datasets. The first goal of our experiments is to show the trade-off between accuracy and fairness measure. To this end, we evaluate our approach with different $\tau$ parameter values and with multiple sensitive attributes. We demonstrate our approach with the Logisitic Regression (LR) classifier and compare it with the standard LR classifier. The second goal is to analyze our approach with multiple sensitive attributes where we show that our approach iteratively removes prejudiced features for each of the known sensitive attributes.  

The final goal of our experiments is to compare the performance metrics of our approach to other known fairness methods. We leverage the fairness-comparison open source library ~\cite{friedler2019comparative} to access datasets and implementations of other known methods. The library provides comparative study of other well-known fairness methods and is meant to facilitate the benchmarking of fairness aware machine learning algorithms. 

\begin{figure*}
	\begin{multicols}{3}
		\begin{subfigure}[b]{1.0\linewidth}
			\includegraphics[width=\linewidth]{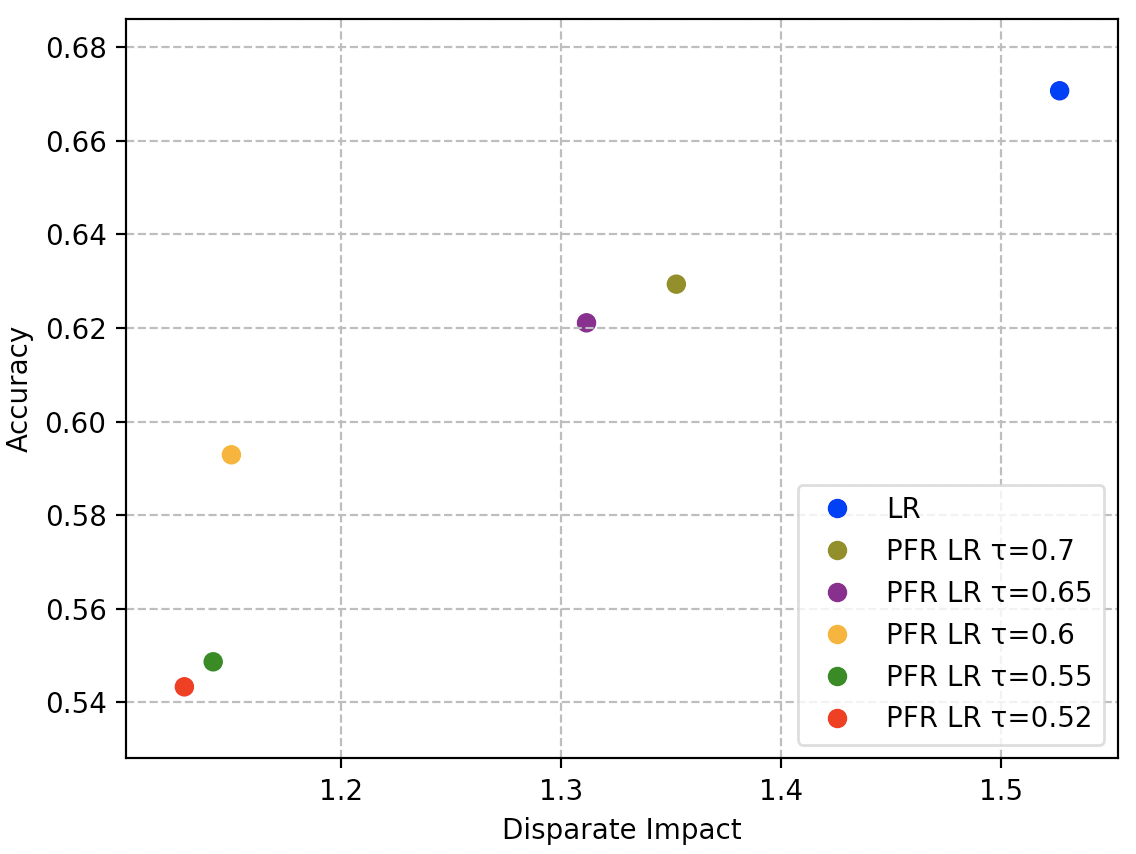}
			\caption{Race Sensitive Attribute }
		\end{subfigure}\par
		\begin{subfigure}[b]{1.015\linewidth}
			\includegraphics[width=\linewidth]{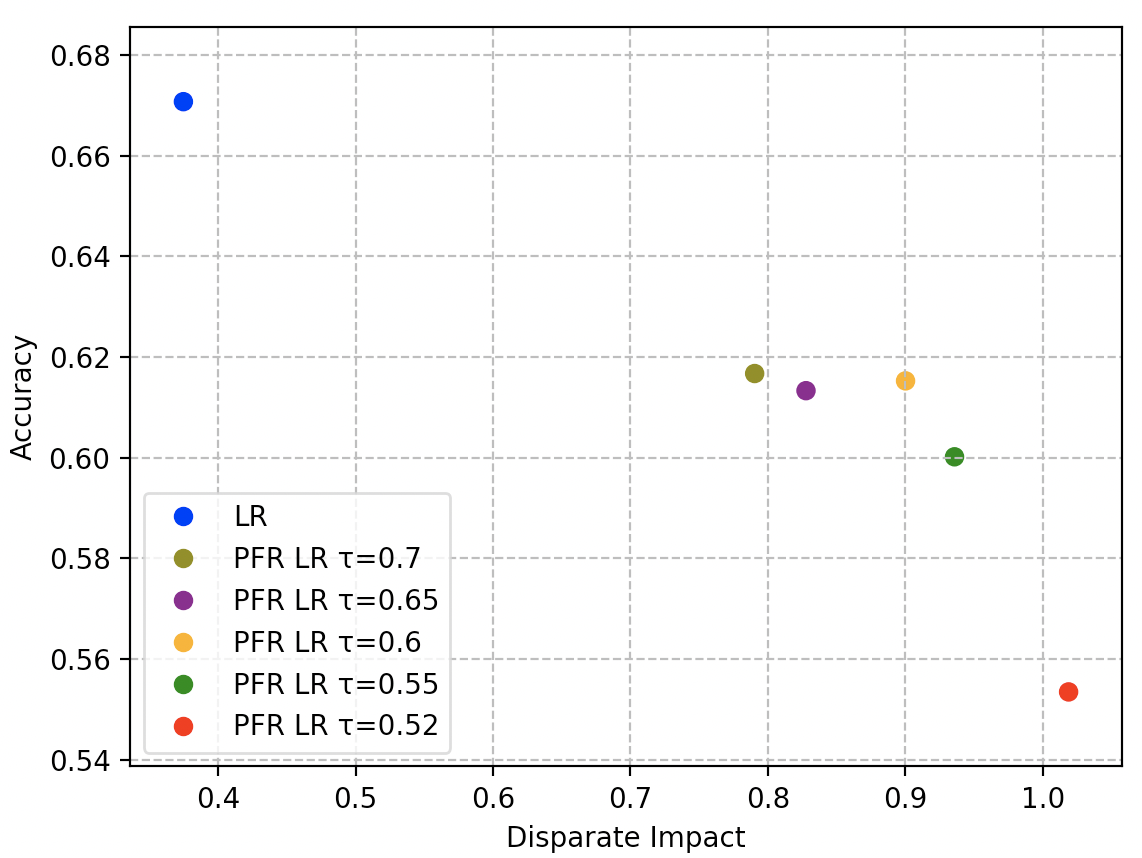}
			\caption{Sex Sensitive Attribute }
		\end{subfigure}\par
		\begin{subfigure}[b]{1.0\linewidth}
			\includegraphics[width=\linewidth]{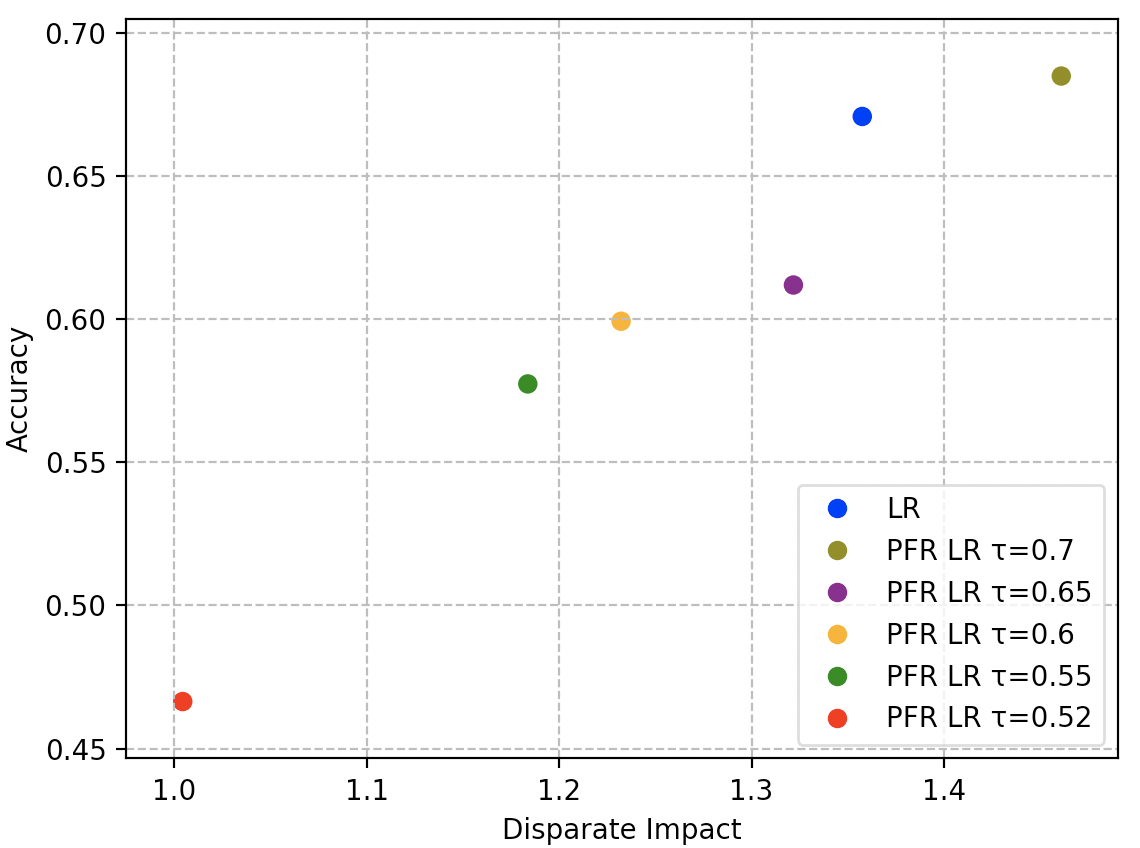}
			\caption{Race \& Sex Sensitive Attributes }
		\end{subfigure}
		
	\end{multicols}
	\caption{Accuracy vs. Disparate Impact evaluation on ProPublica Recidivism Dataset}
	\label{fig:propubpfr}
\end{figure*}

\subsection {Datasets}
\textbf{Adult Income} This dataset contains information about individuals from the 1994 U.S. census. It consists of 32,561 instances and 14 attributes, including sensitive attributes race and sex. The prediction task is determining whether an individual makes more or less than \$50,000 per year ~\cite{friedler2019comparative}. 
\\
\\
\textbf{ProPublica Recidivism} This dataset includes features such as the number of juvenile felonies and the charge degree of the current arrest for 6,167 individuals with 15 attributes, including  sensitive attributes race and sex. This dataset was collected by the
use of the COMPAS risk assessment tool in Broward County, Florida [Angwin et al. 2016]. Each individual has a binary “recidivism” outcome, that is the prediction task, indicating whether they were rearrested within two years after the first arrest ~\cite{friedler2019comparative}. 
\\
\\
For both the datasets, we do minority target classes oversampling since the target classes are skewed in terms of sensitive attributes as target variable. Categorical features in both the datasets are converted to one-hot encoded vectors to make them numerical before passing it to the learning algorithm. We consider \textit{White} and \textit{Male} as privileged classes for Race and Sex sensitive attribute in Adult Income Dataset, and \textit{Caucasian} and \textit{Male} as privileged classes for Race and Sex sensitive attribute in ProPublica Recidivism Dataset respectively.

\begin{table}
	\centering
	\begin{tabular}{ | l | p{5.5 cm} |}
		\hline
		\textbf{Attribute} & \textbf{Values} \\ \hline
		marital-status & Married-AF-spouse, Separated\\ \hline
		native-country & Italy, Haiti, Portugal, Trinadad\&Tobago, Greece, Yugoslavia, Laos, Scotland, Cambodia, Hungary, China, Canada, South,  Poland, Vietnam, Ireland, El-Salvador, Taiwan, Guatemala, Philippines, Honduras, Peru, Hong, Cuba, India, Mexico, Jamaica,Thailand, France, Nicaragua, Columbia, Iran,Germany, United-States \\ \hline
	\end{tabular}
	\caption{PFR Detected Prejudice Inducing Features on Adult Income Dataset with Race Sensitive Attribute at $\tau=0.70$.}
	\label{tab:raceiterative}
\end{table}

\subsection{Accuracy vs. Disparate Impact}
Figure \ref{fig:aidpfr} and \ref{fig:propubpfr} shows accuracy vs. disparate impact trade-off results on Adult Income Dataset and ProPublica Recidivism Dataset respectively. The results show this trade-off on three different set of sensitive attributes i.e. Race, Sex and Race \& Sex, with different $\tau$ parameter values which influences the degree of prejudice removal. As expected, PFR iteratively removes prejudice inducing features from the dataset, as we lower the value of $\tau$, thereby consistently pushing the disparate impact value close to $1.0$, which is the ideal value for a fair model. The figures also show that the trade-off between accuracy and fairness measure can be decided based on different $\tau$ values.

\begin{figure}[h!]
	\begin{subfigure}[b]{0.495\linewidth}
		\captionsetup{justification=centering}
		\includegraphics[width=\linewidth]{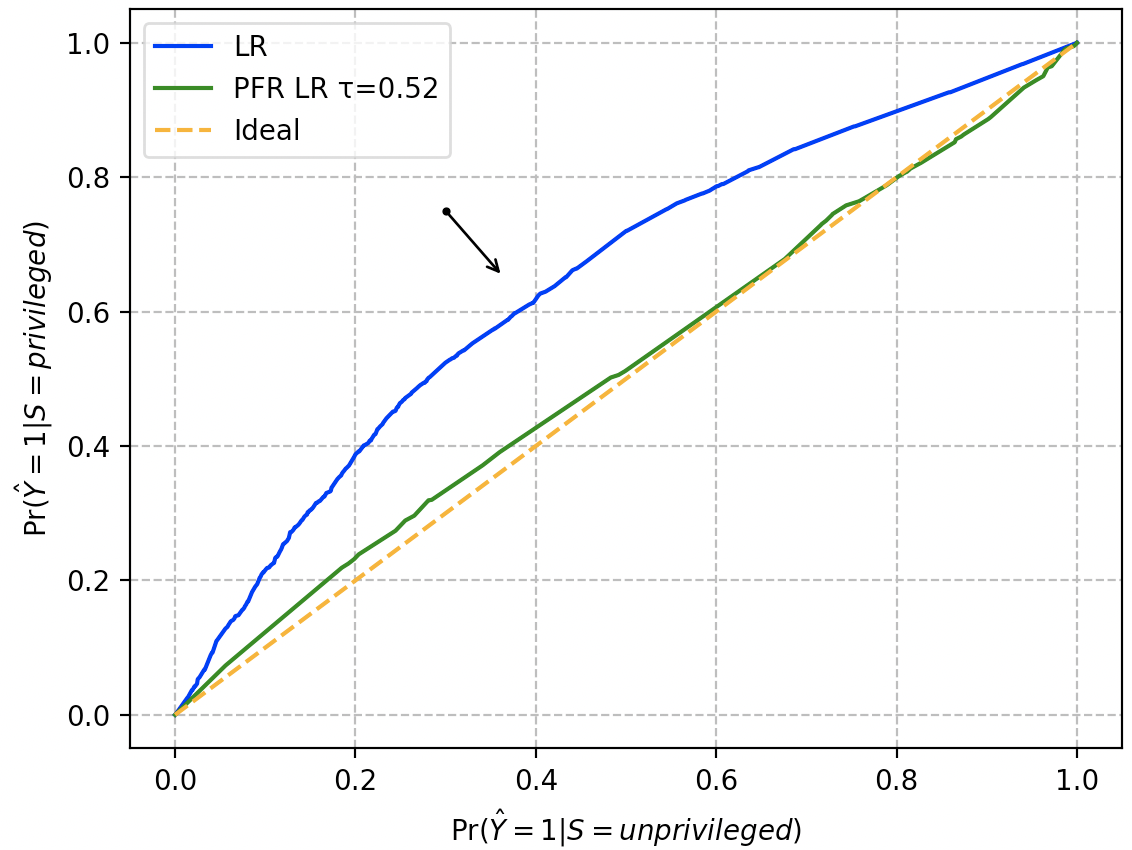}
		\caption{Race Sensitive Attribute}
	\end{subfigure}
	\begin{subfigure}[b]{0.495\linewidth}
		\captionsetup{justification=centering}
		\includegraphics[width=\linewidth]{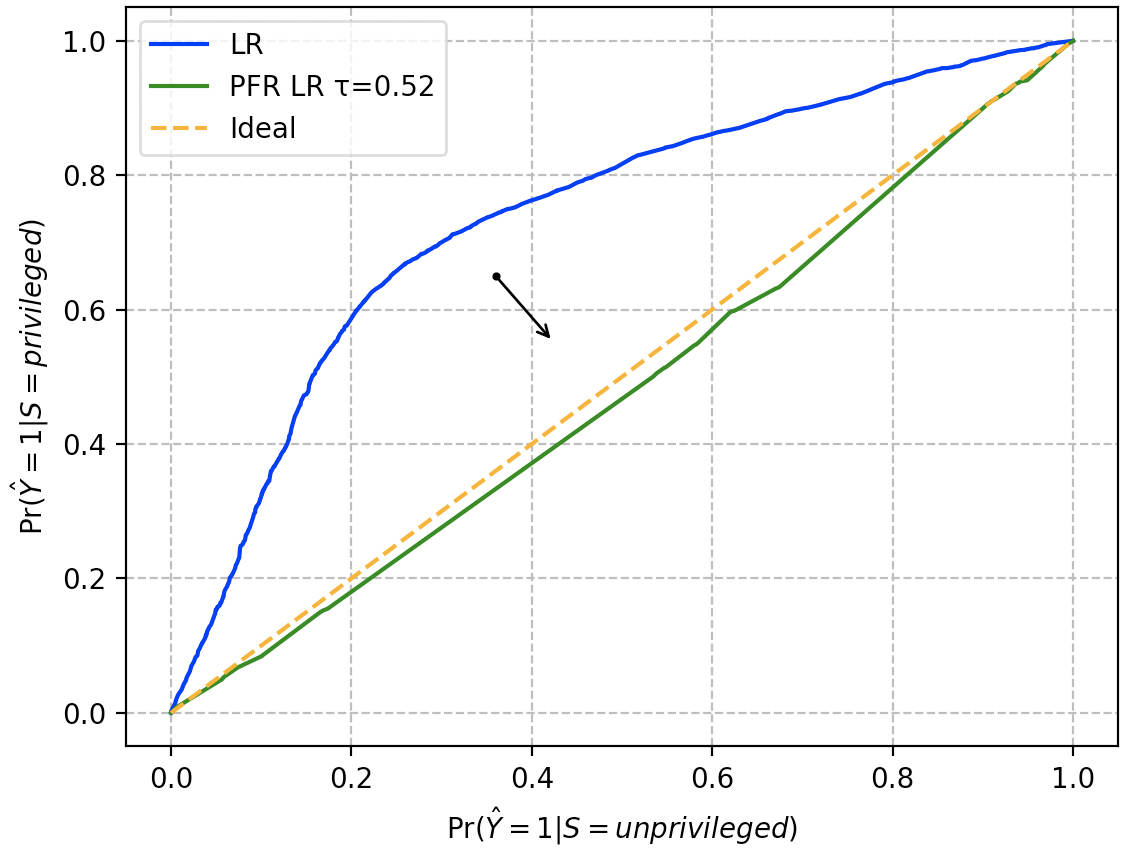}
		\caption{Sex Sensitive Attribute}
	\end{subfigure}
	\caption{PFR push the probability of predicting positive class with respect to privileged and unprivileged classes towards ideal condition on Adult Income Dataset}
	\label{fig:aidpfrcurve}
\end{figure}

Figure \ref{fig:aidpfrcurve} supports this further where we can see that the probability of predicting a positive class with respect to privileged and unprivileged classes is pushed more towards an ideal condition by our approach when compared to the standard LR classifier where ideal condition is the line satisfying positive class predicting probability to be same for both privileged and unprivileged classes.

\begin{table}
	\centering
	\begin{tabular}{ | l | p{5.5 cm} |}
		\hline
		\textbf{Attribute} & \textbf{Values} \\ \hline
		age & continous \\ \hline
		workclass & Self-emp-inc, Without-pay, Local-gov
		\\ \hline
		marital-status & Married-civ-spouse, Divorced, Separated, Widowed, Married-spouse-absent \\ \hline
		hours-per-week & continuous \\ \hline
		native-country & Cambodia, Thailand, Honduras, India, Hungary, Scotland, Dominican-Republic, Outlying-US(Guam-USVI-etc), China \\ \hline
	\end{tabular}
	\caption{PFR Detected Prejudice Inducing Features on Adult Income Dataset with Sex Sensitive Attribute at $\tau=0.70$.}
	\label{tab:sexiterative}
\end{table}

Since PFR iteratively removes the prejudice inducing features, the drop in accuracy of the target model as we lower the value of $\tau$ could be compensated by adding more features that are non-discriminating in nature. Table \ref{tab:raceiterative} and \ref{tab:sexiterative} shows the detected prejudice inducing features at $\tau=0.70$ for Adult Income Dataset with race and sex as sensitive attributes. We can see that with race attribute, the detected prejudice inducing features are mostly about `native-country` which is as expected. In the case of the sex attribute, what is interesting is that ‘hours-per-week‘ is detected as prejudice inducing feature which would not very obvious for human eyes as it takes a continous value and would not have been trivially detected by just looking at the dataset. 

With our approach, we advocate that the disparate impact of near one should be achieved at any cost, and accuracy should be improved only through non-prejudice inducing features. 

\subsection{Multiple Sensitive Attributes}
\begin{figure}[h!]
	\begin{subfigure}[b]{0.495\linewidth}
		\captionsetup{justification=centering}
			\includegraphics[width=\linewidth]{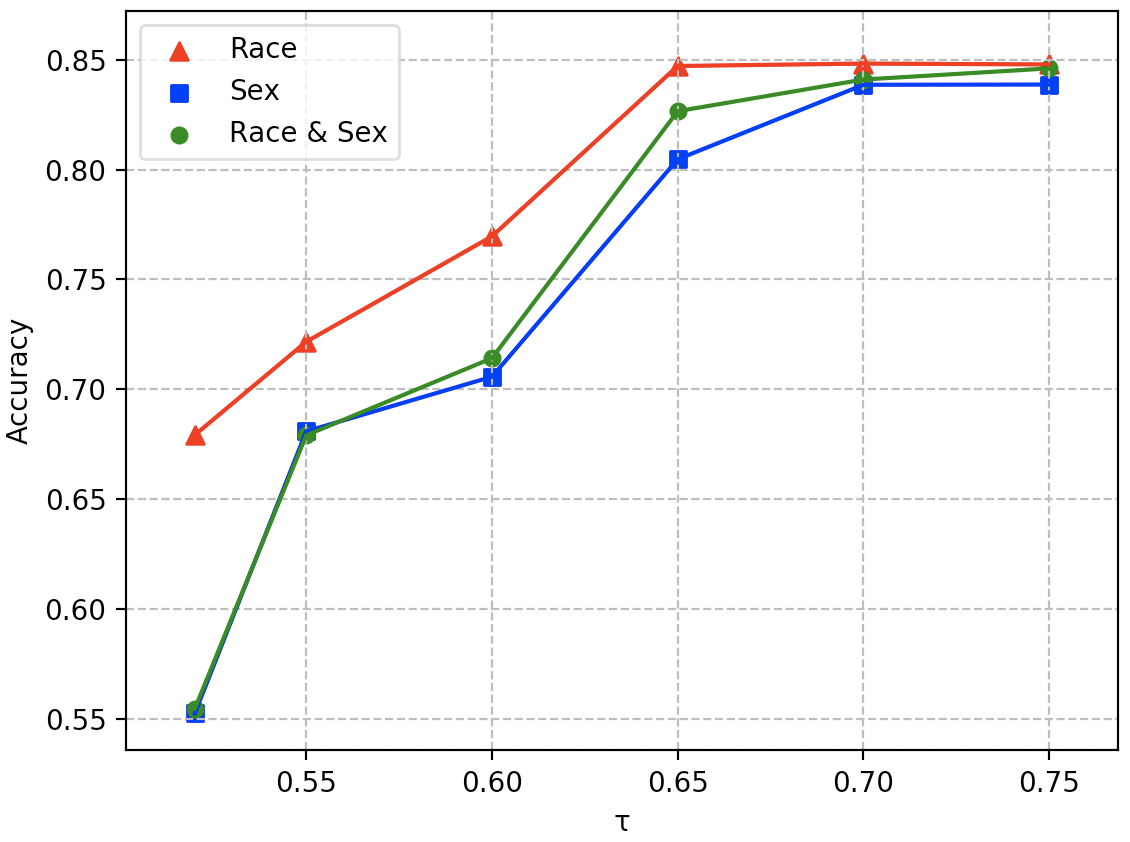}
		\caption{Accuracy}
	\end{subfigure}
	\begin{subfigure}[b]{0.495\linewidth}
		\captionsetup{justification=centering}
		\includegraphics[width=\linewidth]{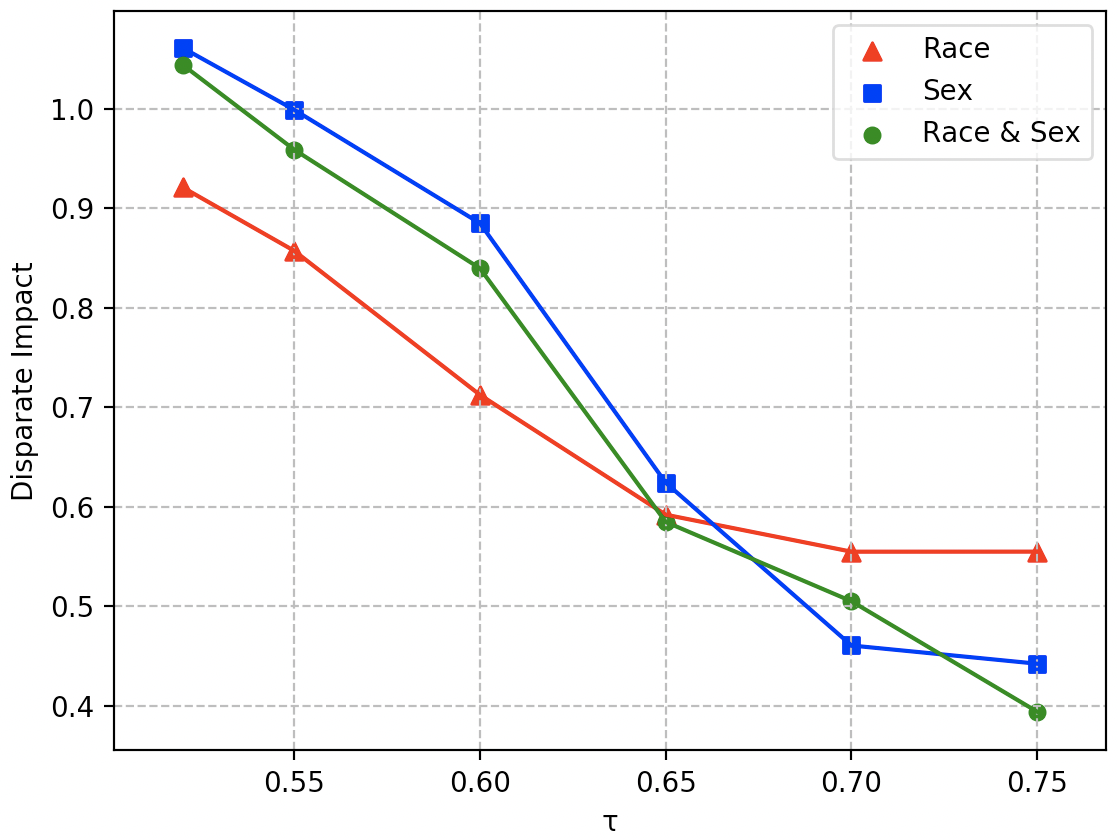}
		\caption{Disparate Impact}
	\end{subfigure}
	\caption{Performance metrics for sensitive attributes at different $\tau$ values on Adult Income Dataset}
	\label{fig:multiple}
\end{figure}

For the tasks where prejudice against multiple sensitive attributes, like race, sex, etc., have to be handled, the proposed approach can be iteratively applied for each of the known sensitive attributes to eliminate prejudice inducing features. Figures \ref{fig:aidpfr}c and \ref{fig:propubpfr}c show working of the approach on Race \& Sex as sensitive attributes. To further understand the relation of applying our approach on multiple attributes, we analyze Accuracy vs. $\tau$ and Disparate Impact vs. $\tau$ performance metrics on Adult Income Dataset. Figure \ref{fig:multiple} shows the result with Race, Sex and Race \& Sex as sensitive attributes. As one can see with Race \& Sex as multiple attributes, our approach results lie somewhere in the middle of Race and Sex as separate sensitive attributes i.e. our approach tries to remove prejudice inducing features in terms of both sensitive attributes, for Accuracy and Disparate Impact case which is also as expected by \cite{friedler2019comparative}.

\subsection {Algorithms Comparison}
We benchmark our algorithm to existing well-known fairness methods; these implementations are leveraged from \cite{friedler2019comparative}.			

In Figure 5a and 5b, we have plotted the output of PFR algorithms at different $\tau$ values and connected the dots to approximately show the trade-off between disparate impact and accuracy as a continuous function though it is not practical to measure the disparate impact metric at all accuracy values. In addition, the output of other well-known algorithms supported by [Friedler et al., 2019] framework are also plotted for comparison. In Figure 5a, we see that PFR beats all except Feldman in terms of disparate impact while giving comparable accuracy. However, in Figure 5b, other algorithms perform marginally better than PFR in terms of accuracy. In both the figures, we could see that PFR outperforms other algorithms in terms of disparate impact metric alone as it could push disparate impact closer to 1.0 at lower $\tau$ value. As previously called out, the accuracy could be improved by adding more features that are not correlated to the sensitive variables. 

\begin{figure}[h!]
	\begin{subfigure}[b]{0.495\linewidth}
		\captionsetup{justification=centering}
		\includegraphics[width=\linewidth]{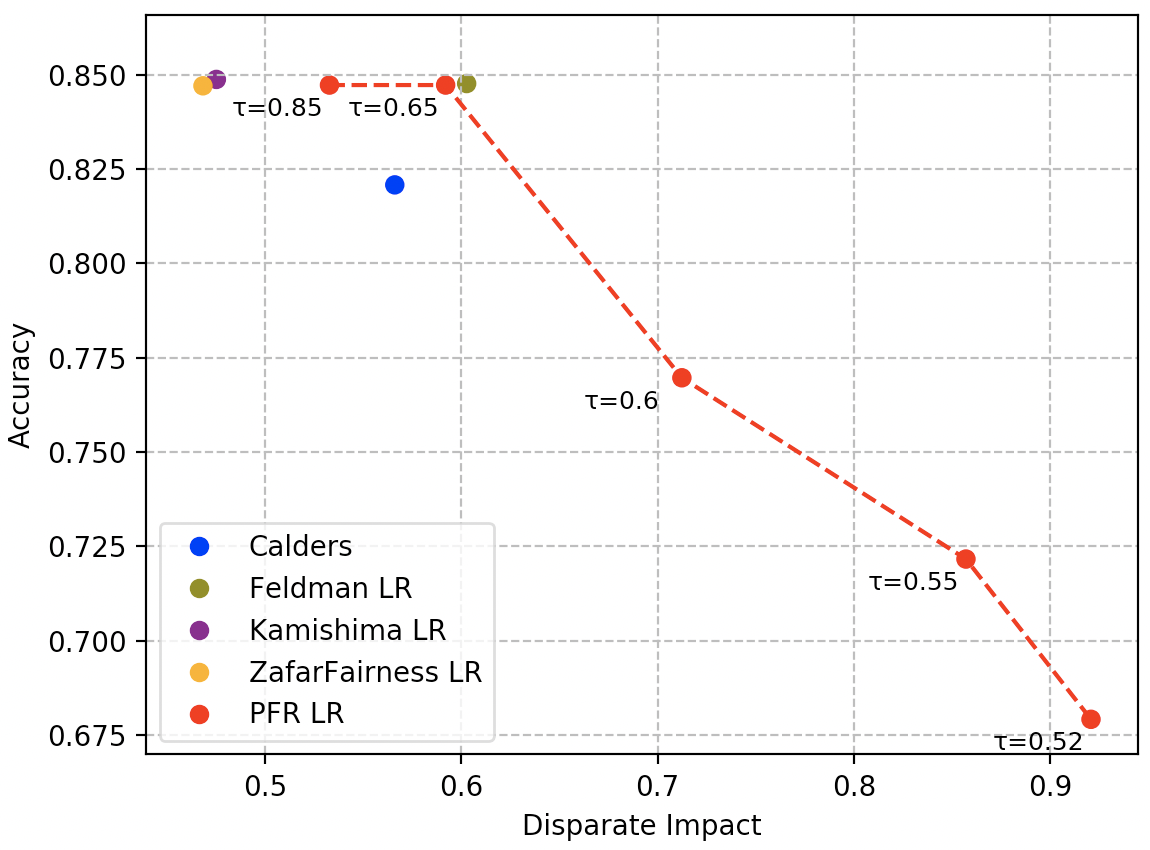}
		\caption{Adult Income \\ Dataset}
	\end{subfigure}
	\begin{subfigure}[b]{0.495\linewidth}
		\captionsetup{justification=centering}
			\includegraphics[width=\linewidth]{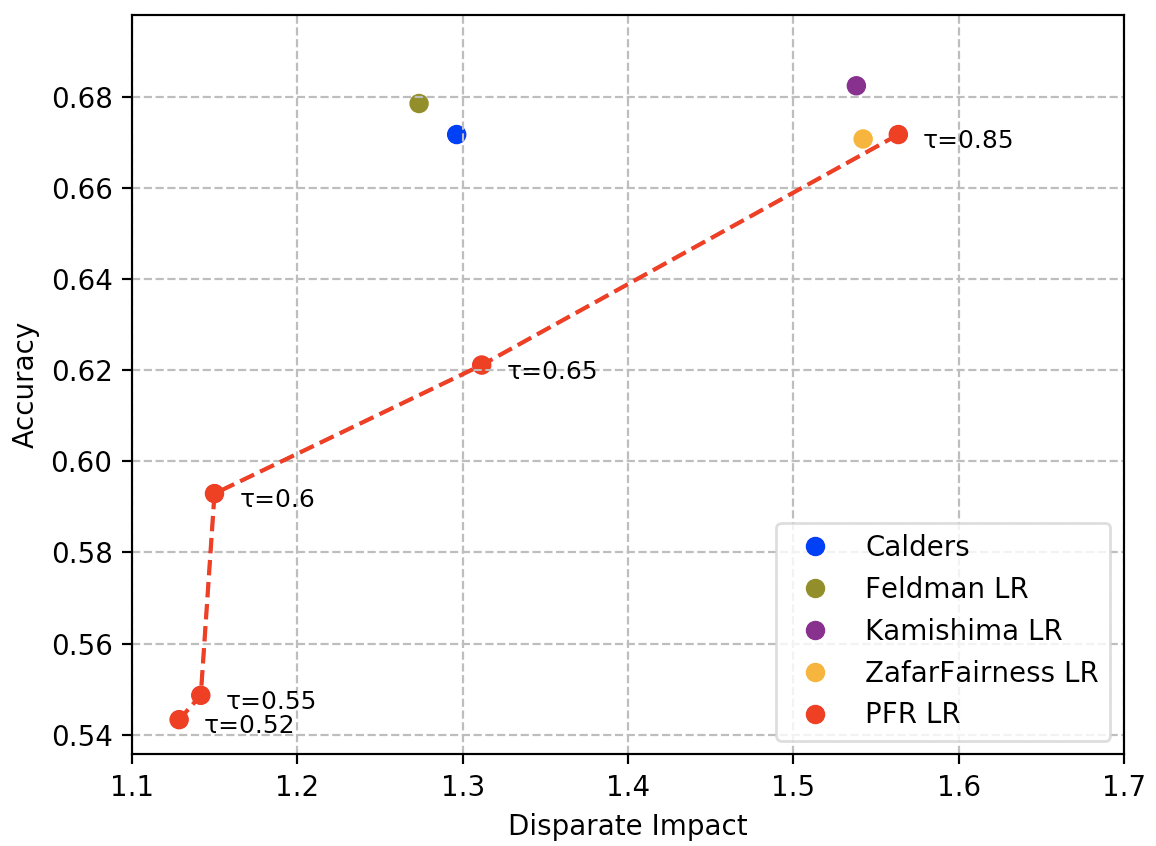}
		\caption{ProPublica Recidivism Dataset}
	\end{subfigure}
	\caption{Accuracy vs. Disparate Impact comparison of differernt algorithms with Race Sensitive Attribute}
	\label{fig:comparison}
\end{figure}

\section{Conclusion \& Future Work}
We proposed an efficient approach that effectively identifies and treats the latent prejudice inducing features that (may) exist in training data. The approach provides a verifiable guarantee that the model is free of prejudice as we use the same algorithm that is used to train the model also to identify and eliminate the prejudice inducing features. Since the approach eliminates prejudice inducing features in an iterative fashion, it provides information about which features are linked to sensitive attributes. The information about the removed features complements the feature engineering process to add more features that are free of correlation with the sensitive attributes, therefore improving accuracy while maintaining fairness measure. Our experiments support that the approach helps to achieve a near-ideal fairness metric.

The approach can be used to identify and remove latent sensitive features and then train a fair model on the residual features using other learning algorithms, both supervised and unsupervised. Also, we could use our approach with in-processing fairness methods by feeding them with information about the detected prejudice inducing features as input. As an extension to the approach, we plan to devise an algorithm that will detect and penalize the prejudice inducing features instead of removing them, by dynamically learning the appropriate penalties for each such feature. 

\bibliographystyle{named}
\bibliography{ijcai20}

\begin{thebibliography}{}

\bibitem[\protect\citeauthoryear{Angwin \bgroup \em et al.\egroup
  }{2016}]{angwin2016machine}
Julia Angwin, Jeff Larson, Surya Mattu, and Lauren Kirchner.
\newblock Machine bias.
\newblock {\em ProPublica, May}, 23:2016, 2016.

\bibitem[\protect\citeauthoryear{Bradley}{1997}]{bradley1997use}
Andrew~P Bradley.
\newblock The use of the area under the roc curve in the evaluation of machine
  learning algorithms.
\newblock {\em Pattern recognition}, 30(7):1145--1159, 1997.

\bibitem[\protect\citeauthoryear{Calders and Verwer}{2010}]{calders2010three}
Toon Calders and Sicco Verwer.
\newblock Three naive bayes approaches for discrimination-free classification.
\newblock {\em Data Mining and Knowledge Discovery}, 21(2):277--292, 2010.

\bibitem[\protect\citeauthoryear{Datta \bgroup \em et al.\egroup
  }{2015}]{datta2015automated}
Amit Datta, Michael~Carl Tschantz, and Anupam Datta.
\newblock Automated experiments on ad privacy settings.
\newblock {\em Proceedings on privacy enhancing technologies}, 2015(1):92--112,
  2015.

\bibitem[\protect\citeauthoryear{Feldman \bgroup \em et al.\egroup
  }{2015}]{feldman2015certifying}
Michael Feldman, Sorelle~A Friedler, John Moeller, Carlos Scheidegger, and
  Suresh Venkatasubramanian.
\newblock Certifying and removing disparate impact.
\newblock In {\em Proceedings of the 21th ACM SIGKDD International Conference
  on Knowledge Discovery and Data Mining}, pages 259--268. ACM, 2015.

\bibitem[\protect\citeauthoryear{Friedler \bgroup \em et al.\egroup
  }{2019}]{friedler2019comparative}
Sorelle~A Friedler, Carlos Scheidegger, Suresh Venkatasubramanian, Sonam
  Choudhary, Evan~P Hamilton, and Derek Roth.
\newblock A comparative study of fairness-enhancing interventions in machine
  learning.
\newblock In {\em Proceedings of the Conference on Fairness, Accountability,
  and Transparency}, pages 329--338. ACM, 2019.

\bibitem[\protect\citeauthoryear{Hardt \bgroup \em et al.\egroup
  }{2016}]{hardt2016equality}
Moritz Hardt, Eric Price, Nati Srebro, et~al.
\newblock Equality of opportunity in supervised learning.
\newblock In {\em Advances in neural information processing systems}, pages
  3315--3323, 2016.

\bibitem[\protect\citeauthoryear{Kamiran and
  Calders}{2009}]{kamiran2009classifying}
Faisal Kamiran and Toon Calders.
\newblock Classifying without discriminating.
\newblock In {\em 2009 2nd International Conference on Computer, Control and
  Communication}, pages 1--6. IEEE, 2009.

\bibitem[\protect\citeauthoryear{Kamishima \bgroup \em et al.\egroup
  }{2012}]{kamishima2012fairness}
Toshihiro Kamishima, Shotaro Akaho, Hideki Asoh, and Jun Sakuma.
\newblock Fairness-aware classifier with prejudice remover regularizer.
\newblock In {\em Joint European Conference on Machine Learning and Knowledge
  Discovery in Databases}, pages 35--50. Springer, 2012.

\bibitem[\protect\citeauthoryear{Sweeney}{2013}]{sweeney2013discrimination}
Latanya Sweeney.
\newblock Discrimination in online ad delivery.
\newblock {\em Queue}, 11(3):10, 2013.

\bibitem[\protect\citeauthoryear{Zafar \bgroup \em et al.\egroup
  }{2017}]{zafar2017fairness}
Muhammad~Bilal Zafar, Isabel Valera, Manuel Gomez~Rodriguez, and Krishna~P
  Gummadi.
\newblock Fairness constraints: Mechanisms for fair classification.
\newblock {\em arXiv preprint arXiv:1507.05259}, 2017.

\bibitem[\protect\citeauthoryear{Zemel \bgroup \em et al.\egroup
  }{2013}]{zemel2013learning}
Rich Zemel, Yu~Wu, Kevin Swersky, Toni Pitassi, and Cynthia Dwork.
\newblock Learning fair representations.
\newblock In {\em International Conference on Machine Learning}, pages
  325--333, 2013.

\bibitem[\protect\citeauthoryear{Zhang and Ntoutsi}{2019}]{zhang2019faht}
Wenbin Zhang and Eirini Ntoutsi.
\newblock Faht: an adaptive fairness-aware decision tree classifier.
\newblock In {\em Proceedings of the 28th International Joint Conference on
  Artificial Intelligence}, pages 1480--1486. AAAI Press, 2019.

\bibitem[\protect\citeauthoryear{Zliobaite}{2015}]{zliobaite2015survey}
Indre Zliobaite.
\newblock A survey on measuring indirect discrimination in machine learning.
  corr abs/1511.00148 (2015).
\newblock {\em arxiv. org/abs/1511.00148}, 2015.

\end{thebibliography}

\end{document}